\documentclass{article}

\usepackage[english]{babel}

\usepackage[letterpaper,top=2cm,bottom=2cm,left=3cm,right=3cm,marginparwidth=1.75cm]{geometry}

\usepackage{amsmath}
\usepackage{graphicx}

\title{Uncovering bias in the PlantVillage dataset}
\author{Mehmet Alican Noyan\\
        \small{Ipsumio B.V.}\\
        \small{alican.noyan@ipsumio.com} \\}

\begin{document}
\maketitle

\begin{abstract}
We report our investigation on the use of the popular “PlantVillage” dataset for training deep-learning based plant disease detection models. We trained a machine learning model using only 8 pixels from the PlantVillage image backgrounds. The model achieved 49.0\% accuracy on the held-out test set, well above the random guessing accuracy of 2.6\%. This result indicates that the PlantVillage dataset contains noise correlated with the labels and deep learning models can easily exploit this bias to make predictions. Possible approaches to alleviate this problem are discussed.
\end{abstract}

\section{Introduction}

Plant diseases are a massive threat to the global food supply. Potato late blight (Phytophthora infestans) caused the Irish Potato Famine that claimed the lives of one million people. Today, more than 150 years later after this catastrophic event, potato late blight costs billions of euros to the global economy each year\cite{Haverkort2008}. Overall, plant diseases are responsible for 20 to 40\% of crop losses globally\cite{Savary2019, Oerke2006}. Disease detection and identification play an essential role in disease management for minimizing crop losses. Researchers have developed various methods for disease detection, such as DNA-based methods and sensing techniques for measuring leaf volatile emissions\cite{Martinelli2015, Li2019}. In 2007, Huan et al. demonstrated one of the first uses of neural networks for plant disease detection\cite{Huang2007}.

The artificial intelligence (AI) revolution started in the early 2010s when convolutional neural networks dominated the computer vision competitions\cite{Ciregan2012, Krizhevsky2012}. The real value of AI, however, was understood when it started tackling challenges in other domains such as medicine\cite{Noyan2020} and physics\cite{Hussain2020}. Today, machine learning has become an indispensable tool in plant science. It has found wide-ranging applications such as classifying plant cell organelles\cite{Li2021}, high-throughput root phenotyping\cite{Lube2022}, and estimating crop growth using drone images\cite{Han2019}. Even though machine learning was used for plant disease identification as early as 2007, a lack of large public datasets prevented further studies. This changed when the first extensive and public plant disease dataset, PlantVillage, was published in 2015\cite{Hughes2015}.

Several open plant disease datasets emerged in recent years (Table \ref{table:1}). The PlantVillage dataset is the largest and most studied plant disease dataset. It contains more than 54,000 images of leaves on a homogenous background. This sparked a plethora of studies on plant disease detection using deep learning\cite{Mohanty2016, Ferentinos2018}. Most of the papers reported classification accuracies above 98\%.

\begin{table}
\centering
\begin{tabular}{c|c|c}
Dataset & Number of images & Google Scholar citations (as of June 2022) \\\hline
PlantVillage\cite{Hughes2015} & 54,305 & 516 \\
BRACOL\cite{Krohling2019} & 4,407 & 11 \\
Plant Pathology\cite{Thapa2020} & 3,651 &33 \\
DiaMOS Plant\cite{Fenu2021} & 3,505 & 1 \\
RoCoLe\cite{Parraga-Alava2019} & 1,560 & 28 \\
Citrus\cite{Rauf2019} & 750 & 77 \\
Rice Leaf Diseases\cite{Prajapati2017} & 120 & 129 \\
\end{tabular}
\caption{Publicly available plant disease datasets}
\label{table:1}
\end{table}

Mohanty et al.\cite{Mohanty2016} realized that the convolution neural network they developed using the PlantVillage dataset experienced a significant drop in accuracy from 99\% to 31\% when tested on other online datasets. They claimed that increasing the number and variability of the dataset would be sufficient to overcome this problem. However, it is also possible that such a vast drop indicates an inherent dataset bias issue. If this is the case, collecting more data will not improve performance. 

Dataset bias sources such as background bias and capture bias are well documented in machine learning\cite{Torralba2011, Tommasi2015}. There is a strong relationship between the environment and the objects in it. For example, an aquatic animal is expected to be found around a body of water. When photographs capture the real world, this effect manifests itself as a correlation between the background and the foreground. Machine learning models may exploit this relationship to predict class labels\cite{Logan2020}. It has been shown that even humans use the background-foreground relationship to make sense of the objects\cite{Torralba2003}. This relationship is called background bias. PlantVillage dataset has a uniform background with no contextual correlation with the foreground (the leaf). Therefore, background bias is not expected in the PlantVillage dataset. In fact, when Mohanty et al.\cite{Mohanty2016} removed the background, the model performance dropped less than 1\%. Another common bias source, capture bias\cite{Torralba2011, Tommasi2015}, emerges from the device used for data collection, collector preferences, and illumination. It affects both the foreground and the background. Therefore, even though background bias is not expected in the PlantVillage dataset, a model can still exploit the background for making predictions due to capture bias. Moreover, removing the background will not be enough since capture bias is also present in the foreground.

This study was performed to investigate and quantify the amount of dataset bias in the most popular public plant disease dataset, PlantVillage. By training a model using a fraction of the background, we were able to show that there is significant capture bias associated with this dataset. We also showed that removing the background does not remove the bias.  This issue should be kept in mind by the current and future users of this and similar datasets.

\section{Methods}
Section 2.1-2.5 describe the datasets used in this paper. Section 2.6 explains the machine learning model development.

\subsection{PlantVillage dataset}

\begin{figure}
\centering
\includegraphics[width=\textwidth]{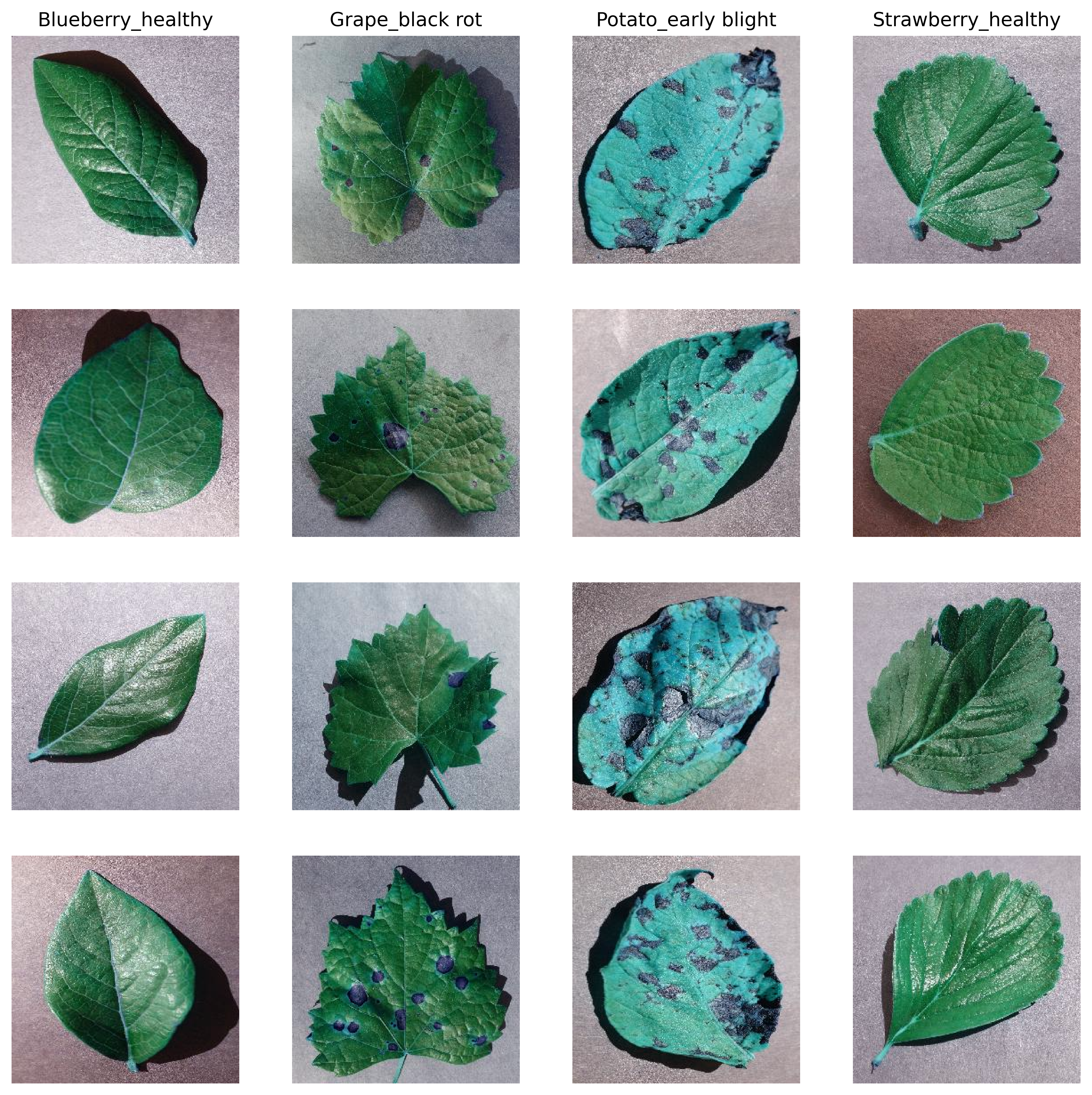}
\caption{\label{fig:PlantVillage}Example images from the PlantVillage dataset. Each column belongs to a single class.}
\end{figure}

The PlantVillage dataset contains 54,305 single leaf images (256 px\textsuperscript{2}, RGB) from 14 crop species (Figure \ref{fig:PlantVillage}). There are 38 classes named as species disease or species healthy (Table \ref{table:2}). The leaves were removed from the plant, placed against a grey or black background, and photographed outdoors with a single digital camera on sunny or cloudy days. Many researchers developed models based on the PlantVillage dataset and reported accuracies above 98\% (Table \ref{table:3}). In this paper, we used the version of the dataset shared by Mohanty et al., which contains the original images, grayscale images and images without the backgrounds\cite{Mohanty2016}.

\begin{table}[h]
\centering
\begin{tabular}{c|c|c}
Apple\_apple scab & Grape\_leaf blight & Strawberry\_leaf scorch\\
Apple\_blackrot & Grape\_healthy & Strawberry\_healthy\\
Apple\_cedar apple rust	& Orange\_huanglongbing	& Tomato\_bacterial spot\\
Apple\_healthy	& Peach\_bacterial spot	& Tomato\_early blight\\
Blueberry\_healthy	& Peach\_healthy	& Tomato\_late blight\\
Cherry\_powdery mildew	& Bell Pepper\_bacterial spot	& Tomato\_leaf mold\\
Cherry\_healthy	& Bell Pepper\_healthy	& Tomato\_septoria leaf spot\\
Corn\_gray leaf spot	& Potato\_early blight	& Tomato\_spider mites\\
Corn\_common rust	& Potato\_late blight	& Tomato\_target spot\\
Corn\_northern leaf blight	& Potato\_healthy	& Tomato\_tomato yellow leaf curl virus\\
 Corn\_healthy	& Raspberry\_healthy	& Tomato\_tomato mosaic virus\\
Grape\_black rot	& Soybean\_healthy	& Tomato\_healthy\\
Grape\_esca	& Squash\_powdery mildew\\
\end{tabular}
\caption{38 class names of the PlantVillage dataset.}
\label{table:2}
\end{table}

\begin{table}[h]
\centering
\begin{tabular}{c|c|c|c}
Reference & Dataset & Plant & Accuracy\\\hline
Mohanty et al.\cite{Mohanty2016}& PlantVillage & All & 99.35\%\\
Too et al.\cite{Too2019}& PlantVillage & All & 99.75\%\\
Ferentinos et al.\cite{Ferentinos2018} & PlantVillage extension & All & 99.53\%\\
Zhang et al.\cite{Zhang2018} & PlantVillage extension & Maize & 98.9\%\\
\end{tabular}
\caption{Example neural network performances for the PlantVillage and its extensions.}
\label{table:3}
\end{table}

\subsection{PlantVillage extensions}

A common criticism\cite{Fenu2021} of the PlantVillage dataset is that a single leaf on a black/gray background does not represent an image taken in the field (Figure \ref{fig:Cukurova}). Therefore, some researchers extended the PlantVillage dataset with field data or images collected from other online databases (Table \ref{table:3}). However, adding new images does not guarantee to eliminate the bias; on the contrary, it can introduce additional biases.

\subsection{PlantVillage\_8px}

\begin{figure}[h]
\centering
\includegraphics[width=0.8\textwidth]{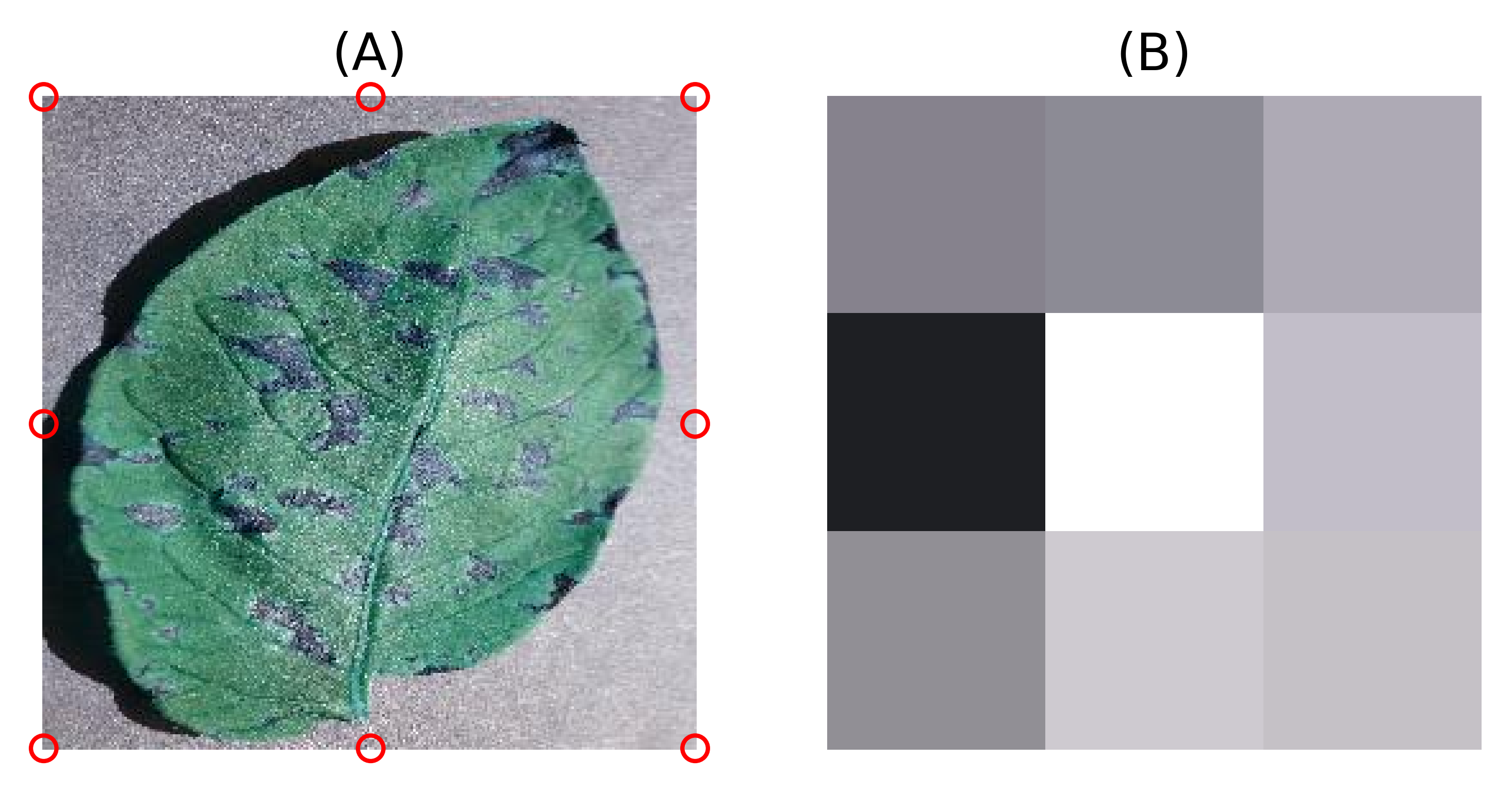}
\caption{\label{fig:Transformation}Transformation of a single image into 8 pixels. (a) The original image with red circles showing pixel locations (b) 8 pixels reduced to a square}
\end{figure}

\begin{figure}[h]
\centering
\includegraphics[width=0.5\textwidth]{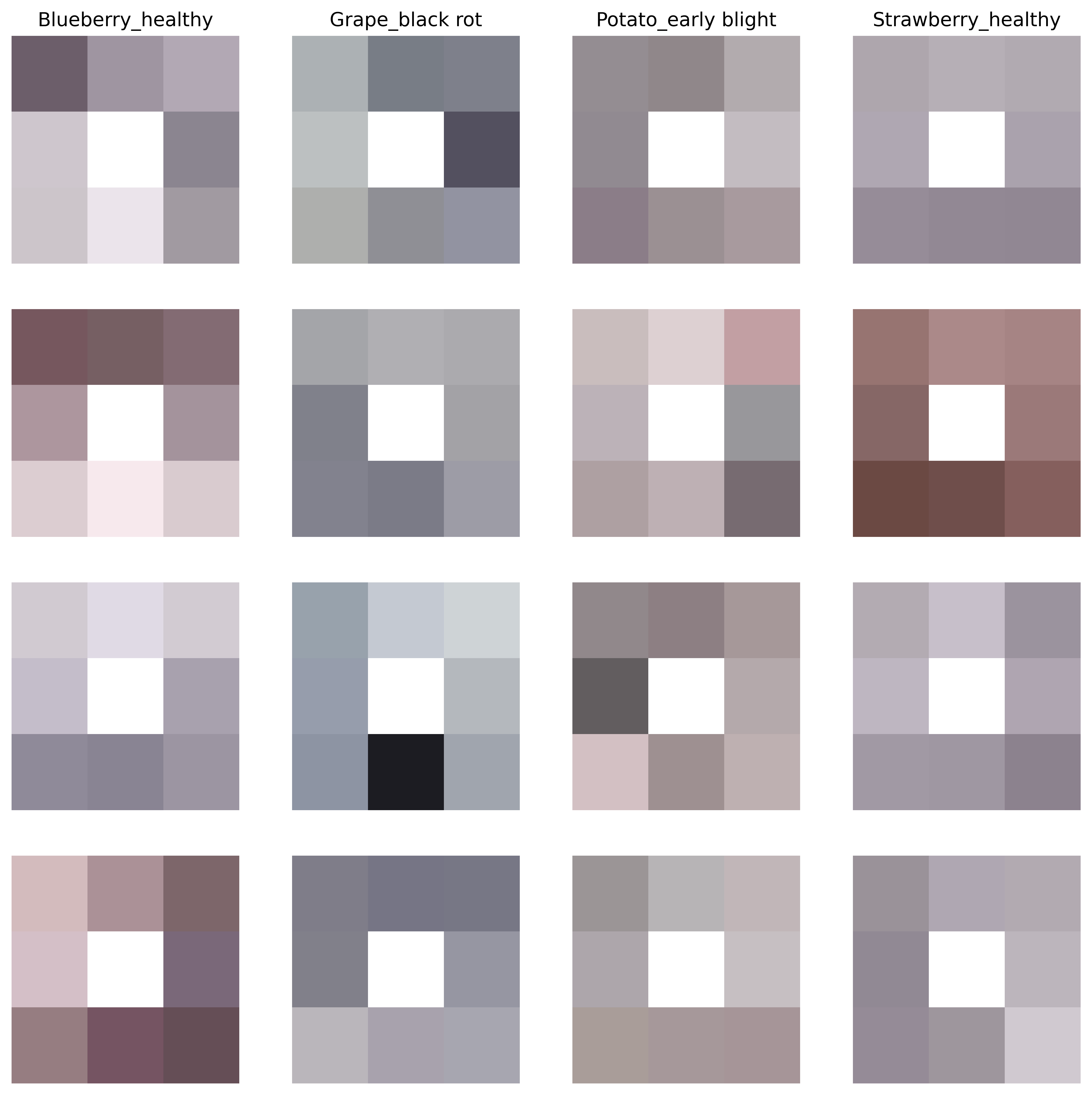}
\caption{\label{fig:PlantVillage8px}Example images from the PlantVillage\_8px dataset. Original images can be seen in Figure \ref{fig:PlantVillage}}
\end{figure}

There are multiple ways to separate the background and the foreground\cite{Logan2020}. We devised a transformation that uses a fraction of the background pixels. It extracts 8 pixels from the original image, four from the corners and four from the centers of the sides. An example transformation can be seen in Figure \ref{fig:Transformation}. This transformation was applied to all images of the PlantVillage dataset (Figure \ref{fig:PlantVillage}) to create the PlantVillage\_8px dataset (Figure \ref{fig:PlantVillage8px}). As can be seen, PlantVillage\_8px contains zero information about diseases (the signal). It only includes information on the background color and capture conditions, which is pure noise in terms of plant diseases. Moreover, 8 pixels correspond to less than 0.1\% of the whole background.

\subsection{MNIST\_8px}

MNIST is a dataset containing 28 px\textsuperscript{2} grayscale handwritten digit images (Figure \ref{fig:MNIST}). Digits are centered, and the backgrounds are uniform. The 8-pixel transformation described in Figure 3 was applied to MNIST, resulting in the MNIST\_8px dataset. The background in this dataset does not contain background bias or capture bias; therefore, it will be used as an ideal case comparison.

\subsection{PlantVillage\_blur}

Blur is an image quality metric that is part of the capture conditions. It is expected to be similar across the classes. In order to quantify the effect of background removal on capture bias, we used blur as a predictor before and after background removal. The data repository contained the PlantVillage dataset and the foregrounds (PlantVillage\_fg). By subtracting the foregrounds from the PlantVillage dataset, we obtained the backgrounds as well (PlantVillage\_bg). We measured the blur of every image in all these three datasets and obtained PlantVillage\_blur, PlantVillage\_blur\_fg, and PlantVillage\_blur\_bg. A summary of PlantVillage datasets can be found in Table \ref{table:datasetsummary}.

\begin{table}[h]
\centering
\begin{tabular}{c|c|c}
Dataset name & Description & Feature size\\\hline
PlantVillage & Original dataset & 256x256\\
PlantVillage\_fg & Foregrounds of PlantVillage & 256x256\\
PlantVillage\_bg & Backgrounds of PlantVillage & 256x256\\
PlantVillage\_8px & 8 pixels from the PlantVillage & 8\\
PlantVillage\_blur & Blur values of the PlantVillage images  & 1\\
PlantVillage\_fg\_blur & Blur values of the PlantVillage\_fg images & 1\\
PlantVillage\_bg\_blur & Blur values of the PlantVillage\_bg images & 1\\

\end{tabular}
\caption{Summary of the PlantVillage dataset transformations}
\label{table:datasetsummary}
\end{table}

\subsection{Machine learning model development}

In order to quantify the amount of bias in the PlantVillage dataset, we trained and tested a machine learning model on the PlantVillage\_8px dataset. If there is no bias, the model should not be able to beat the random guess accuracy, which is defined as 100/number\_of\_classes \% for a balanced dataset.

Then, to show that removing the backgrounds does not necessarily remove this bias, we trained and tested another three machine learning models on the datasets PlantVillage\_blur, PlantVillage\_fg\_blur, and PlantVillage\_bg\_blur. If background removal indeed helps, the bias contained in the PlantVillage\_fg\_blur should be much less than PlantVillage\_blur and PlantVillage\_bg\_blur.

We used scikit-learn's random forest classifier implementation with default hyperparameters to train all four models\cite{Pedregosa2011}. To be comparable with other works that developed models on PlantVillage, all four datasets were randomly split into a training set (80\%) and a test set (20\%). Classification accuracy was used to evaluate model performance.

\section{Results and Discussion}

Table \ref{table:4} shows the result of our model on the PlantVillage\_8px dataset. It reached 49.0\% classification accuracy on the test set, well above random guess performance (100/38 = 2.6 \%). On the MNIST\_8px dataset, the same model obtained 11.7\% accuracy, which is roughly equal to the random guess performance (100/10 = 10\%).

These results indicate significant dataset bias in the PlantVillage dataset. Since the foregrounds and the backgrounds are not correlated contextually, minimal background bias is expected. Therefore, capture bias must be the main reason for the dataset bias. This means that the models developed on this dataset will experience significant performance drops even on new datasets collected on similar conditions, let alone field data. Note that this underestimates the dataset bias because capture bias influences both the background and foreground, whereas the model used only a fraction of the background. Moreover, the random forest model was trained with the default hyperparameters without any tuning to improve its performance.

\begin{table}[h]
\centering
\begin{tabular}{c|c|c}
Dataset name & Random guess accuracy & Random Forest Model\\\hline
PlantVillage\_8px & 2.6\% & 49.0\%\\
MNIST\_8px & 10\% & 11.7\%\\

\end{tabular}
\caption{Random forest performance on PlantVillage\_8px and MNIST\_8px datasets}
\label{table:4}
\end{table}

To eliminate this bias, Mohanty et al.\cite{Mohanty2016} removed the backgrounds and trained a model with only foregrounds. However, our experiment shows that background removal does not eliminate the bias, PlantVillage\_fg\_blur contains an amount of bias similar to the datasets with backgrounds (Table \ref{table:5}). Capture bias (in this case blur) influences both the foreground and background.

\begin{table}[h]
\centering
\begin{tabular}{c|c|c|c}
PlantVillage\_blur & PlantVillage\_fg\_blur & PlantVillage\_bg\_blur & Random guess accuracy\\\hline
11.7\% & 10.0\% & 10.8\% & 2.6\%\\
\end{tabular}
\caption{The effectiveness of background removal on avoiding capture bias.}
\label{table:5}
\end{table}

Extending this dataset with new images can indeed alleviate the issue, but only if great care is taken. One of the most famous studies with an extended PlantVillage dataset is reported by Ferentinos et al.\cite{Ferentinos2018}. The dataset is not open, but the collection procedure described in the paper suggests that it introduces yet another bias. They extended the classes from 38 to 58. Unfortunately, 31\% of the classes were purely field data, whereas 48\% of the classes were collected purely from the laboratory (Table \ref{table:6}). Therefore, the model will learn the difference between the field images and laboratory images to distinguish between these classes easily.

\begin{table}[h]
\centering
\begin{tabular}{c|c}
Source & Number of classes (\%)\\\hline
Field conditions & 18 (31\%)\\
Laboratory conditions & 28 (48\%)\\
Mixed & 12 (21\%)\\
\end{tabular}
\caption{Data sources from Ferentinos et al.\cite{Ferentinos2018}}
\label{table:6}
\end{table}

At the end of the day, the best way to deal with a biased dataset is to avoid collecting it in the first place. Design of Experiments, a branch of statistics, laid out the principles of efficient and proper data collection\cite{Montgomery2019}. The salient idea is to determine the noise factors before data collection and ensure they are either controlled for or randomized. If one must work with a biased dataset, the first step is to understand the bias sources and quantify them. Once this is done, bias can be decreased by either removing it or negating it with additional data collection. The most critical step is to collect a separate dataset that matches the use case and report the model performance on this dataset. This will provide a reliable estimate of the model performance.

\section{Conclusions}

This work identified and quantified the dataset bias in the PlantVillage dataset. Unfortunately, removing backgrounds or haphazardly adding data cannot eliminate this bias. Therefore, researchers should be careful when using this dataset and the models developed on this dataset for their future research and applications.

\bibliographystyle{ieeetr}
\bibliography{PlantVillage}

\newpage
\renewcommand{\thefigure}{S\arabic{figure}}
\setcounter{figure}{0}

\section{Supplementary Information}

\begin{figure}[h]
\centering
\includegraphics[width=0.4\textwidth]{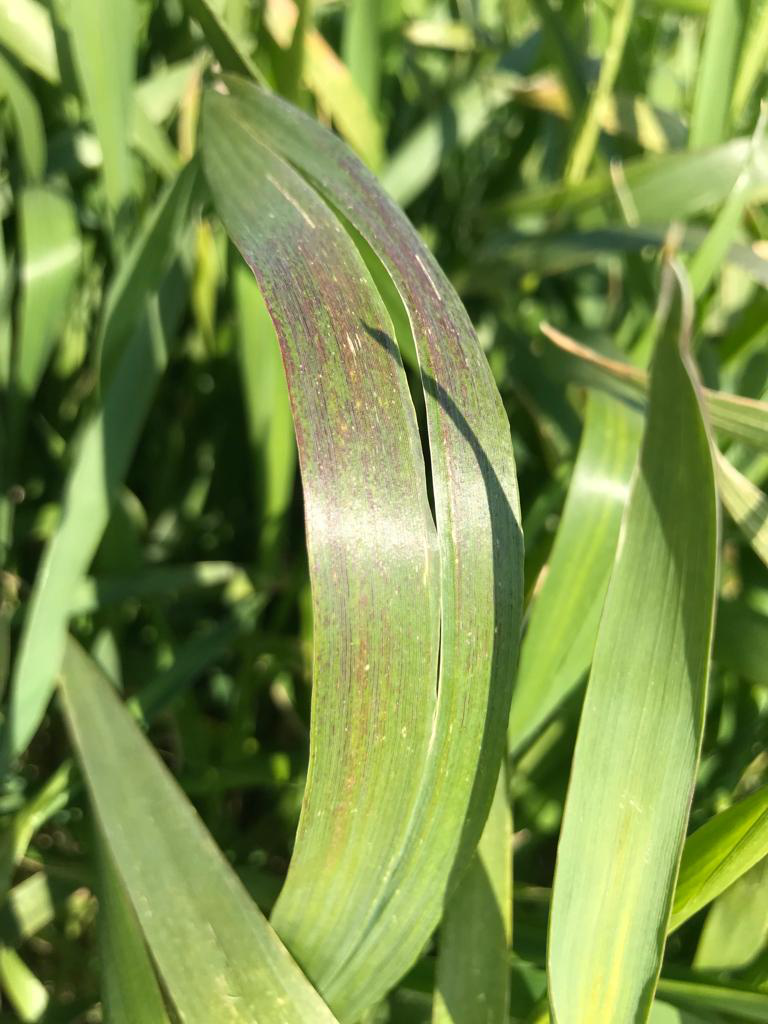}
\caption{\label{fig:Cukurova}Wheat leaf image from the Cukurova region.}

\end{figure}

\begin{figure}[h]
\centering
\includegraphics[width=0.4\textwidth]{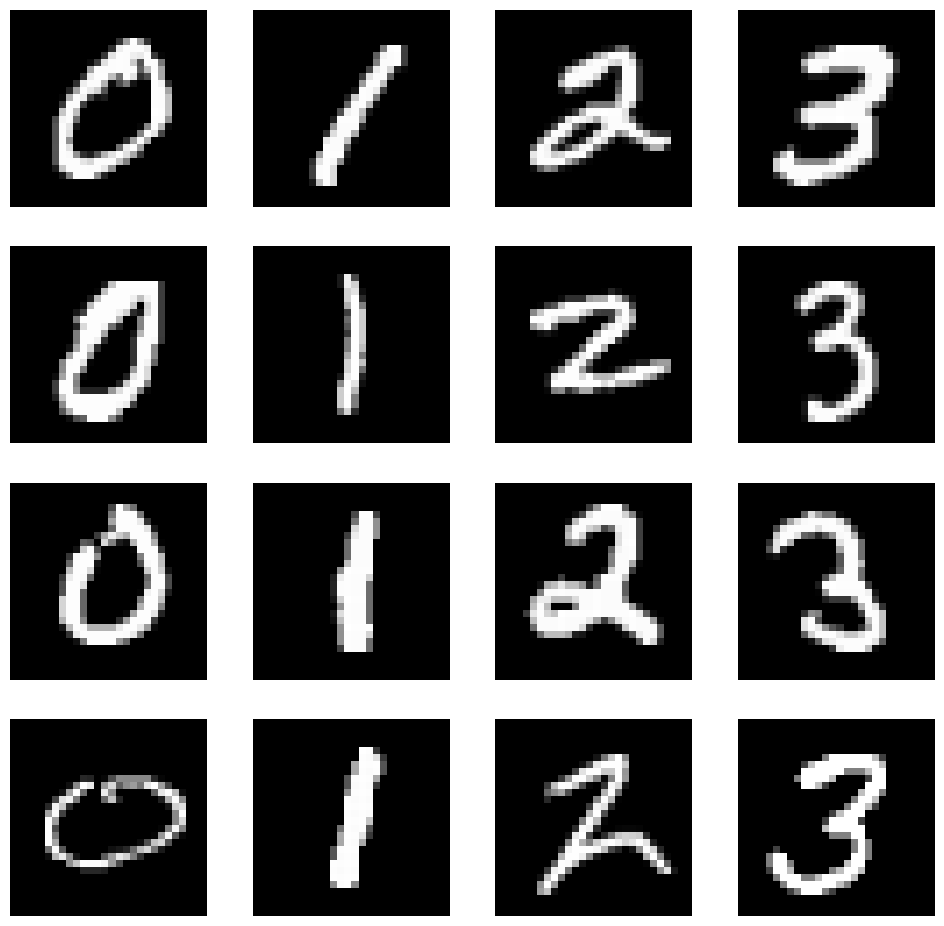}
\caption{\label{fig:MNIST}MNIST handwritten digits dataset.}
\end{figure}

\end{document}